\documentclass{article}
\usepackage{arxiv}
\usepackage[utf8]{inputenc} 
\usepackage[T1]{fontenc}    
\usepackage{hyperref}       
\usepackage{url}            
\usepackage{booktabs}       
\usepackage{amsfonts}       
\usepackage{nicefrac}       
\usepackage{microtype}      
\usepackage{lipsum}
\usepackage{algorithm}
\usepackage{algorithmic}
\usepackage{graphicx}
\usepackage{comment}
\usepackage{wrapfig}
\usepackage{multirow}
\usepackage{makecell}
\usepackage{color}
\usepackage{amsmath} 
\usepackage{amssymb}           
\usepackage{amsfonts}            
\usepackage{mathrsfs}
\usepackage{amsfonts,amssymb,amsmath,amsthm}
\newtheorem{theorem}{Theorem}

\newtheorem{lemma}{Lemma}

\newtheorem{proposition}[theorem]{Proposition}
\newtheorem{definition}{Definition}

\newtheorem{example}{Example}

\newcommand{\squishlist}{
\begin{list}{{{\small{$\bullet$}}}}
{\setlength{\itemsep}{3pt}      \setlength{\parsep}{1pt}
\setlength{\topsep}{1pt}       \setlength{\partopsep}{0pt}
\setlength{\leftmargin}{1em} \setlength{\labelwidth}{1em}
\setlength{\labelsep}{0.5em} } }
\newcommand{\squishend}{  \end{list}  }

\usepackage[]{mathtools} 
\def\##1\#{\begin{align}#1\end{align}}
\def\$#1\${\begin{align*}#1\end{align*}}
\usepackage{enumitem}
\usepackage[]{amsthm} 
\usepackage[]{amssymb}
\usepackage{dsfont}
\usepackage{hyperref}
\usepackage{url}
\usepackage{xcolor}
\usepackage{amsmath,amsbsy}

\usepackage{tcolorbox}
\definecolor{grey}{rgb}{0.33, 0.33, 0.33}

\newcommand{\p}{\mathbb{P}}

\newcommand{\E}{\mathbb E}

\newcommand{\kr}{\textsf{Kr}}

\newcommand{\nY}{\tilde{Y}}
\newcommand{\dnn}{\text{DNN}}

\graphicspath{ {./images/} }

\title{Identifiability of Label Noise Transition Matrix}

\author{Yang Liu$^1$, Hao Cheng$^1$, Kun Zhang$^{2,3}$\\ 
	$^1$ University of California, Santa Cruz \\					
	$^2$ Carnegie Mellon University \\
	$^3$ Mohamed bin Zayed University of Artificial Intelligence\\
	{\tt \{yangliu, haocheng\}@ucsc.edu, kunz1@cmu.edu}
}

\begin{document}
	
	\maketitle
	
	\begin{abstract}
		The noise transition matrix plays a central role in the problem of learning with noisy labels. Among many other reasons, a large number of existing solutions rely on access to it. Identifying and estimating the transition matrix without ground truth labels is a critical and challenging task.  When label noise transition depends on each instance, the problem of identifying the instance-dependent noise transition matrix becomes substantially more challenging. Despite recent works proposing solutions for learning from instance-dependent noisy labels, the field lacks a unified understanding of when such a problem remains identifiable. The goal of this paper is to characterize the identifiability of the label noise transition matrix.  Building on Kruskal's identifiability results, we are able to show the necessity of multiple noisy labels in identifying the noise transition matrix for the generic case at the instance level. We further instantiate the results to explain the successes of the state-of-the-art solutions and how additional assumptions alleviated the requirement of multiple noisy labels. Our result also reveals that disentangled features are helpful in the above identification task and we provide empirical evidence.  
	\end{abstract}
	
	\section{Introduction}

The literature of learning with noisy labels concerns the scenario when the observed labels $\nY$ can differ from the true one $Y$.
The noise transition matrix $T(X)$, defined as the transition probability from $Y$ to $\nY$ given $X$, plays a central role in this problem. 
Among many other benefits, the knowledge of $T(X)$ has demonstrated its use in performing either risk \cite{natarajan2013learning,Patrini_2017_CVPR}, or label \cite{Patrini_2017_CVPR}, or constraint corrections \cite{wang2021fair}. In beyond, it also finds applications in ranking small loss samples \cite{han2020sigua} and detecting corrupted samples \cite{zhu2021good}. On the other hand, applying the wrong transition matrix $T(X)$ can lead to a number of issues. The literature has well-documented evidence that a wrongly inferred transition matrix can lead to performance drops \cite{natarajan2013learning,liu2021can,xia2019anchor,zhu2021clusterability}, and false sense of fairness \cite{wang2021fair,liu2021can}. Knowing whether a $T(X)$ is identifiable or not helps understand if the underlying noisy learning problem is indeed learnable.

The earlier results  have focused on class- but not instance-dependent transition matrix $T(X) \equiv T:=[\p(\nY =j  | Y=i)]_{i,j}, \forall X$. The literature has provided discussions of the identifiability of $T$ under the mixture proportion estimation setup \cite{scott2015rate}, and has identified a reducibility condition for inferring the inverse noise rate. Later works have developed a sequence of solutions to estimate $T$ under a variety of assumptions, including irreducibility \cite{scott2015rate}, anchor points \cite{liu2016classification,xia2019anchor,yao2020dual}, separability \cite{cheng2017learning}, rankability \cite{northcutt2017learning,northcutt2021confident}, redundant labels \cite{liu2018surrogate}, clusterability \cite{zhu2021clusterability}, among others \cite{zhang2021learning,li2021provably}.

The question of identifying and estimating $T$ becomes much trickier when the noise transition matrix is instance-dependent. The potentially complicated dependency between $X$ and $T(X)$ renders it even less clear whether solving this problem is viable or not. We observe a recent surge of different solutions towards solving the instance-dependent label noise problem \cite{cheng2017learning,xia2020parts,cheng2020learning,yao2021instance}. Some of the results took on the problem of estimating $T(X)$, while the others proposed solutions to learn directly from instance-dependent noisy labels. We will survey these results in Section \ref{sec:related}.

Despite the above successes, there lacks a unified understanding of when this learning from instance-dependent noisy label problem is indeed identifiable and therefore learnable.
The mixture of different observations calls for the need for demystifying: 
(1) Under what conditions are the noise transition matrices $T(X)$ identifiable? (2) When and why do the existing solutions work when handling the instance-dependent label noise? (3)  When $T(X)$ is not identifiable, what can we do to improve its identifiability? Providing answers to these questions will be the primary focus of this paper. The main contributions of this paper are to characterize the identifiability of instance-dependent label noise, use them to explain the current existing results and point out possible directions to improve. Among other findings, some highlights of the paper are 1. We find many existing solutions have a deep connection to the celebrated Kruskal's identifiability results that date back to the 1970s \cite{kruskal1976more,kruskal1977three}. 2. Three separate independent and identically distributed (i.i.d.) noisy labels (random variables) are both necessary and sufficient for instance-level identifiability. 3. Disentangled features help with identifiability.

Our paper will proceed as follows. Section \ref{sec:formulation} and \ref{sec:pre} will present our formulation and the highly relevant preliminaries. Section \ref{sec:instance} provides characterizations of the identifiability at the instance level and lays the foundations for our discussions. Section \ref{sec:population} extends the discussion to different instantiations that help us explain the sucess of existing solutions. Section \ref{sec:exp} provides some empirical observations and Section \ref{sec:conclude} concludes this paper.

\subsection{Related works}\label{sec:related}

In the literature of learning with label noise, a major set of works focus on designing \emph{risk-consistent} methods, i.e., performing empirical risk minimization (ERM) with specially designed loss functions on noisy distributions leads to the same minimizer as if performing ERM over the corresponding unobservable clean distribution.
The \emph{noise transition matrix} is a crucial component for implementing risk-consistent methods, e.g., loss correction \cite{patrini2017making}, loss reweighting \cite{liu2015classification}, label correction \cite{xiao2015learning} and unbiased loss \cite{natarajan2013learning}. A number of solutions were proposed to estimate this transition matrix for class-dependent label noise, which we have discussed in the introduction. 

To handle instance-dependent noise, recent solutions include estimating local transition matrices for different groups of data \cite{xia2020parts}, using confidence scores to revise transition matrices \cite{berthon2020confidence}, and using clusterability of the data \cite{zhu2021clusterability}. More recent works have used the causal knowledge to improve the estimation \cite{yao2021instance}, and the deep neural network to estimate the transition matrix defined between the noisy label and the Bayes optimal label \cite{yang2021estimating}. Other works chose to focus on the learning from instance-dependent label noise directly, without explicitly estimating the transition matrix \cite{zhu2020second, cheng2020learning,berthon2021confidence,xia2020robust,Li2020DivideMix}.

The identifiability issue with label noise has been discussed in the literature, despite not being formally treated. Relevant to us is the identifiability results studied in the Mixture Proportion Estimation setting \cite{scott2015rate,yao2020towards,menon2015learning}. We'd like to note that the identifiability was defined for the inverse noise rate, which differs from our focus on the noise transition matrix $T$. 
To our best knowledge, we are not aware of other works that specifically address the identifiability of $T(X)$, particularly for an instance-dependent label noise setting. Highly relevant to us is the Kruskal's identifiability results \cite{kruskal1976more,kruskal1977three,sidiropoulos2000uniqueness,allman2009identifiability}, which reveals a sufficient condition for identifying a parametric model that links a hidden variable to a set of observed ones. Some of this paper's efforts include translating the Kruskal's and its follow-up results to the problem of learning with noisy labels. 
	
\section{Formulation}\label{sec:formulation}

We will use $(X,Y)$ to denote a supervised data in the form of (feature, label) drawn from an unknown distribution over $X \times Y$. We consider a $K$-class classification problem where the label $Y \in \{1,2,...,K\}$ with $K \geq 2$. In our setup, we do not observe the clean true label $Y$, but rather a noisy one, denoting by $\tilde{Y}$. The generation of $\tilde{Y}$ follows the following transition matrix:
$
    T(X) := [\p(\tilde{Y} = j|Y=i,X)]_{i,j=1}^K
$.
That is $T(X)$ is a $K \times K$ matrix with its $(i,j)$ entry being $\p(\tilde{Y} = j|Y=i,X)$.

To define identifiability, we will denote by $\Omega$ an observation space. We first define identifiability for a general parametric space $\Theta$. Denote the distribution induced by the parameter $\theta \in \Theta$ of a statistical model on the observation space $\Omega$ as $\p_{\theta}$ \cite{kruskal1976more,allman2009identifiability}. To give an example, for a fixed $X$ (when consider instance-level identifiability), and $\Omega$ is simply the outcome space for its associated noisy label $\tilde{Y}$, i.e., $\{1,2,...,K\}$. In this case, each $\theta$ is the combination of a possible transition matrix $T(X)$ and the hidden prior of $\p(Y|X)$, which we use to denote the conditional probability distribution of $Y$ given $X$. $\p_{\theta}$ is then the distribution (probability density function) $\p(\tilde{Y}|X)$. Later in Section \ref{sec:instance} when we introduce three noisy labels $\tilde{Y}_1,\tilde{Y}_2,\tilde{Y}_3$ for each $X$,  $\p_{\theta}$ is the joint distribution $\p(\tilde{Y}_1,\tilde{Y}_2,\tilde{Y}_3|X)$.
Identifiability defines as follows:
\begin{definition}[Identifiability] The parameter $\theta$ (statistical model) is identifiable if $\p_{\theta} \neq \p_{\theta'}, \forall \theta \neq \theta'$.
\end{definition}
We now formally define identifiability for the task of learning with noisy labels for an instance $X$. Denote by $\theta(X):=\{T(X),\p(Y|X)\}$. $\p_{\theta(X)}$ is the distribution (probability density function) over $\Omega$, defined by the noise transition matrix $T(X)$ and the prior $\p(Y|X)$. To emphasize, $\Omega$ is not necessarily the observation space of the noisy label $\nY$ only. The exploration of an effective $\Omega$ will be one of the focuses of the paper. 
\begin{definition}
[Identifiability of $T(X)$] 
For a given $X$, $T(X)$ is identifiable if $\p_{\theta(X)} \neq \p_{\theta'(X)}$ for $\theta(X) \neq \theta'(X)$, up to label permutation. 
\end{definition}
Label permutation relabels the label space, e.g., $1 \rightarrow 2,~2 \rightarrow 1$, and the rows in $T(X)$ will swap.


\section{Preliminary}\label{sec:pre}

In this section, we will introduce two highly relevant results on Mixture Proportion Estimation (MPE) \cite{scott2015rate} and Kruskal's identifiability result \cite{kruskal1976more,kruskal1977three}. 

\subsection{Preliminary results using irreducibility and anchor points}
The problem of learning from noisy labels ties closely to another problem called Mixture Proportion Estimation (MPE) \cite{scott2015rate}, which concerns the following problem: let $F,J,H$ be distributions defined over a Hilbert space $\mathcal Z$. The three relate to each other as follows:
$
F = (1-\kappa^*)J + \kappa^* H
$.
The identifiability problem concerns the ability to identify the mixture proportion $\kappa^*$ from only observing $F$ and $H$. The following identifiability result has been established:
\begin{proposition}\cite{blanchard2010semi}
$\kappa^*$ is identifiable if $J$ is irreducible with respect to $H$, that $J$ can not be written as $J = \gamma H + (1-\gamma) F'$, where $0 \leq \gamma \leq 1$, and $F'$ is another distribution. 
\end{proposition}

Later, the anchor point condition \cite{yao2020towards}, a stronger requirement of the irreducibility was established:
\begin{proposition} \cite{yao2020towards}
$\kappa^*$ is identifiable if there exists a subset $S \subseteq \mathcal Z$ such that $H(S) > 0$, but $\frac{J(S)}{H(S)} = 0$, where $J(S), H(S)$ denote the probabilities of $S$ measured by $J,H$.
\end{proposition}
 The above set $S$ is called an anchor set. 
A sequence of follow-up works have emphasized the necessity of anchor points in identifying a class-dependent noise transition matrix $T$ \cite{xia2019anchor,li2021provably}. 

Prior work has established the connection between the MPE problem and the learning from noisy label one \cite{yao2020towards} for the identifiability of an inverse noise rate $\p(Y|\nY)$ but not the noise transition $T(X)$. We reproduce the discussion and fill in the gap. 
The discussion and results are for the class-dependent but not instance-dependent label noise, i.e., $T(X) \equiv T$, and for a binary classification problem. To follow the convention, we assume $Y \in \{-1,+1\}$. There are two things we need to do:
(1) State the noisy label problem as an MPE one; and (2) show that the identifiability of $\kappa^*$ is equivalent to the identifiability of $T$.
We start with the first thing above. We want to acknowledge that this equivalence appeared before in \cite{yao2020towards,menon2015learning}. 
We reproduce it here to make our paper self-contained.
We first present:
\begin{lemma}\label{lemma:mpe}
Denote by  $\tilde{\pi}_- = \frac{\pi_-}{1-\pi_+}, \tilde{\pi}_+= \frac{\pi_+}{1-\pi_-}$ and we have
\begin{align}
  \p(X|\tilde{Y}=-1) &= \tilde{\pi}_- \cdot \p(X|\tilde{Y}=+1) + (1-\tilde{\pi}_-)\cdot \p(X|Y=-1)\\
    \p(X|\tilde{Y}=+1) &= \tilde{\pi}_+ \cdot \p(X|\tilde{Y}=-1) + (1-\tilde{\pi}_+)\cdot \p(X|Y=+1)~.
\end{align}
\end{lemma}
 Now $ \p(X|\tilde{Y}=+1), \p(X|\tilde{Y}=-1)$ correspond to the observed mixture distribution $F,H$, while $\p(X|Y=+1)$ and $\p(X|Y=-1)$ are the two unobserved $J$s, $\tilde{\pi}_-, \tilde{\pi}_+$ correspond to the mixture proportion $\kappa^*$. This has established the learning with noisy label problem as two MPE problems corresponding for the two associated distributions $\p(X|\tilde{Y}=-1), \p(X|\tilde{Y}=+1)$.  Therefore to formally establish the equivalence between identifying  $\kappa^*$ and $T$, we will only need to establish the equivalence between identifying $\tilde{\pi}_-, \tilde{\pi}_+$ and identifying $T$.
 Denote by $ e_+:=\p(\nY=-1|Y=+1), e_- := \p(\nY=+1|Y=-1)$ which determine the $T$ for the binary case.
%
We have the following equivalence theorem:
\begin{theorem}
Identifying $\{\tilde{\pi}_-, \tilde{\pi}_+\}$ is equivalent with identifying  $\{e_-,e_+\}$. \label{thm:equiv}
\end{theorem}
The above theorem concludes the same irreducibility and anchor point conditions proposed under MPE also apply to identifying noise transition matrix $T$. This conclusion aligns with previous successes in estimating class-dependent noise transition matrix $T$ when the anchor point conditions are satisfied \cite{liu2016classification,xia2019anchor,li2021provably}. The above result has \textbf{limitations}. Notably, the result focuses on two mixed distributions, leading to the binary classification setup in the noisy learning setting. The authors did not find an easy extension to the multi-class classification problem.  Secondly, the translation to the noisy learning problem requires the noise transition matrix to stay the same for a distribution of $X$ (e.g., $\p(X|\nY=+1)$), instead of providing instance-level understanding for each $X$.

\subsection{Kruskal's identifiability result}

Our results build on the Kruskal's identifiability result \cite{kruskal1976more,kruskal1977three}. The setup is as follows: suppose that there is an unobserved variable $Z$ that takes values in a $K$-sized discrete domain $\{1,2,...,K\}$. $Z$ has a non-degenerate prior $ \p(Z = i) > 0$. Instead of observing $Z$, we observe $p$ variables $\{O_i\}_{i=1}^p$. Each $O_i$ has a finite state space $\{1,2,...,\kappa_i\}$ with cardinality $\kappa_i$. Let $M_i$ be a matrix of size $K \times \kappa_i$, which $j$-th row is simply $[\p(O_i = 1|Z=j),...,\p(O_i = \kappa_i|Z=j)]$. In this case, $[M_1,M_2,...,M_p]$ and $\p(Z = i)$ are the hidden parameters that control the generation of observations - together, these form our $\theta$. We now introduce the Kruskal rank of a matrix, which plays a central role in Kruskal's identifiability results. 
\begin{definition}[Kruskal rank] \cite{kruskal1976more,kruskal1977three}
For a matrix $M$, the Kruskal rank of $M$ is the largest number $I$ such that every set of $I$ rows \footnote{There exists other definition that checks columns. Results would be symmetrical.} of $M$ are linearly independent. 
\end{definition}
In this paper, we will use $\kr(M)$ to denote the Kruskal rank of matrix $M$. To give an example, 
$M = \begin{small}
\begin{bmatrix}
1 & 0 & 0\\
0 & 1 & 0\\
2 & 0 & 0 \\
\end{bmatrix}
\end{small}\Rightarrow \kr(M) = 1
$.
This is because $[1, 0,0]$ and $[2,0,0]$ are linearly dependent. We first reproduce the following theorem:
\begin{theorem}\cite{kruskal1976more,kruskal1977three,sidiropoulos2000uniqueness}
The parameters $M_i, i=1,...,p$ are identifiable, up to label permutation, if
\vspace{-0.1in}
\begin{align}
    \sum_{i=1}^p \emph{\kr}(M_i) \geq 2K+p-1  \label{eqn:Kruskal}
\end{align}
\label{thm:Kruskal}
\end{theorem}
\vspace{-0.15in}
The result for $p=3$ was first established in \cite{kruskal1977three}, and then it was shown in \cite{sidiropoulos2000uniqueness} that the proof extends to a general $p$. The proof builds on showing that different parameter $\theta$ leads to different stacking of $M$s: $[M_1,...,M_p]$. For example, when $p=3$, $[M_1,M_2,M_3]$ forms the tensor of the observations.

\section{Instance-Level Identifiability}\label{sec:instance}


This section will characterize the identifiability of $T(X)$ at the instance level. 

\subsection{Single noisy label might not be sufficient and setting up for multiple noisy labels}
At a first sight, it is impossible to identify $\p(\tilde{Y}|Y,X) $ from only observing $\p(\tilde{Y}|X)$,\footnote{We clarify that we will require knowing $\p(\tilde{Y}|X)$ - this requirement may appear weird when only one noisy label is sampled. But in practice, there are tools available to regress the posterior function $\p(\tilde{Y}|X)$ for each $X$.} unless $X$ satisfies the anchor point definition that $\p(Y = k|X) =1$ for a certain $k$: since
$
    \p(\tilde{Y}|X) = \p(\tilde{Y}|Y,X) \cdot \p(Y|X)
$, different combinations of $\p(\tilde{Y}|Y,X), \p(Y|X)$ can lead to the same $ \p(\tilde{Y}|X) $. More specifically, consider the following example:
\begin{example}Suppose we have a binary classification problem with
$
T(X) = \begin{bmatrix}
1-e_-(X) & e_-(X) \\
e_+(X) & 1-e_+(X)
\end{bmatrix}
$. Note that using chain rule we have
\begin{align*}
\p(\tilde{Y}=+1|X) &= \p(\tilde{Y}=+1|Y=+1,X) \cdot \p(Y=+1|X)  +\p(\tilde{Y}=+1|Y=-1,X) \cdot \p(Y=-1|X)\\
& =(1-e_+(X)) \cdot \p(Y=+1|X) + e_-(X) \cdot \p(Y=-1|X)
\end{align*}
Consider two cases:
$
\text{(1):~}\p(Y=+1|X)=1,~ e_+(X) = e_-(X) = 0.3 ~~\text{and}~~\text{(2):~}\p(Y=+1|X) = 0.7,~ e_+(X) = 0.1, ~e_-(X) = 0.233.
$
Both cases will return the same $\p(\tilde{Y}=+1|X) = 0.7$.
\end{example}

Is then the anchor point requirement necessary for identifying $T(X)$ at the instance level? 
The discussion in the rest of this section departs from the classical single noisy label setting. Instead, we assume for each instance $X$, we will have $p$ conditionally independent (given $X,Y$) and identically distributed noisy labels $\tilde{Y}_1,...,\tilde{Y}_p$ generated according to $T(X)$. Let's assume for now we potentially have these labels. Later in this section, we discuss when having multiple redundant labels are possible, and connect to existing solutions in the literature in the next section.

Before we formally present the results for having multiple conditionally independent noisy labels, we offer intuitions. The reason behind this identifiability result ties close to latent class model \cite{clogg1995latent} and tensor decomposition \cite{anandkumar2014tensor}. When the $p$ noisy labels are conditionally independent given $X$ and $Y$, we will have the joint distribution written as:
$
\p(\nY_1,\nY_2,...,\nY_p|Y, X) = \prod_{i=1}^p \p(\nY_i|Y, X)
$
That is, the joint distribution of noisy labels can be encoded in a much smaller parameter space! In our setup, when we assume the i.i.d. $\nY_i, i=1,2,...,p$ are generated according to the same transition matrix $T(X)$, the parameter space is fixed and determined by the size of $T(X)$. Yet, when we increase $p$, the observation space $\p(\nY_1,\nY_2,...,\nY_p|Y,X)$ becomes richer to help us identify $T(X)$.

\subsection{The necessity of multiple noisy labels}

We first define an \emph{informative noisy label}.
\begin{definition}\label{def:rank}
For a given $(X,Y)$, we call their noisy label $\tilde{Y}$ informative if $\text{rank}(T(X)) = K$. 
\end{definition}
Definition \ref{def:rank} simply requires the rank of $T(X)$ to be full, which is already assumed in the literature - e.g., loss correction \cite{natarajan2013learning,Patrini_2017_CVPR} would require the matrix has an inverse $T^{-1}(X)$, which is equivalent to $T(X)$ being full rank. In particular, it was required $e_+(X) + e_-(X) < 1$ in \cite{natarajan2013learning}, which can be easily shown to imply $T(X)$ is full rank. 
Our first identifiability result states as follows:

\begin{theorem}\label{thm:main:instance}
With i.i.d. noisy labels, three informative noisy labels $\tilde{Y}_1,\tilde{Y}_2,\tilde{Y}_3$ ($p=3$) are both sufficient and necessary to identify $T(X)$.
\end{theorem}
\begin{proof} [Proof sketch] We provide the key steps of the proof. The full proof can be found in the supplemental material. 
We first prove sufficiency. We first relate our problem setting to the setup of Kruskal's identifiability scenario: $Y \in \{1,2,...,K\}$ corresponds to the unobserved hidden variable $Z$. $\p(Y = i)$ corresponds to the prior of this hidden variable. Each $\nY_i, i=1,...,p$ corresponds to the observation $O_i$. $\kappa_i$ is then simply the cardinality of the noisy label space, $K$. In the context of this theorem, $p=3$, corresponds to the three noisy labels we have. 

Each $\nY_i$ corresponds to an observation matrix $M_i$:
$
M_i[j,k] = \p(O_i = k|Z=j) = \p(\nY_i = k|Y=j, X)
$. Therefore, by definition of $M_1,M_2,M_3$ and $T(X)$, they all equal to $T(X)$: $M_i \equiv T(X), i = 1,2,3$. When $T(X)$ has full rank, we know immediately that all rows in $M_1,M_2,M_3$ are independent. Therefore, the Kruskal ranks satisfy
$
\kr(M_1) = \kr(M_2) = \kr(M_3) = K
$. Checking the condition in Theorem \ref{thm:Kruskal}, we easily verify
\begin{align*}
  \kr(M_1) + \kr(M_2) + \kr(M_3) = 3K \geq 2K+2  ~.
\end{align*}
Calling Theorem \ref{thm:Kruskal} proves the sufficiency. 

To prove necessity, we need to prove less than 3 informative labels will not suffice to guarantee identifiability. The idea is to show that the two different sets of parameters $T(X)$ can lead to the same joint distribution $\p(\nY_1,\nY_2|
X)$. We leave the detailed constructions to the supplemental material. 

\end{proof}

The above result points out that to ensure identifiability of $T(X)$ at the instance level, we would need three conditionally independent and informative noisy labels. This result coincides with a couple of recent works that promote the use of three redundant labels \cite{liu2018surrogate,zhu2021clusterability}. Per our theorem, these two proposed solutions have a more profound connection to the identifiability of hidden parametric models, and three labels are not only algorithmically sufficiently, but also necessary.

 The crowdsourcing community has been largely focusing on soliciting more than one label from crowdsourced workers, yet the learning from noisy label literature has primarily focused on learning from a single one. One of the primary motivations of crowdsourcing multiple noisy labels is indeed to aggregate them into a cleaner one \cite{liu2012variational,shah:nips13,sig15}, which serves as a pre-processing step towards solving the noisy learning problem. Nonetheless, our result demonstrates the other significance of having multiple labels - they help the learner identify the underlying true noise transition parameters. There have been discussions about crowdsourcing additional label information being unnecessary when we have robust learning strategies in hand. This section presents a strong case against the above argument.

\section{Instantiations and Practical Implications of Our Identifiability Results}\label{sec:population}

Most of the learning with noisy label solutions focuses on the case of using a single label and have observed empirical successes. 
In this section, we provide extensions of our results to cover of state-of-the-art learning with noisy label methods, together with specific assumptions over $X$, $T(X) = [\p(\tilde{Y}|Y,X)]$ etc. We show that our results can easily extend to these specific instantiations that successfully avoided the requirements of having multiple noisy labels for each $X$. The high-level intuition for Section \ref{sec:clusterability} is to leverage the smoothness and clusterability of the nearest neighbor $X$s so that their noisy labels will jointly serve as the multiple noisy labels for the local group. Section \ref{sec:group1} and \ref{sec:group2} build on the notion that if $T(X)$ is the same for a group of $X$s, each group can then be treated as one ``instance" and a ``disentangled" version of $X$ will become observation variables that serve the similar role of the additionally required noisy labels. 


\subsection{Leveraging smoothness and clusterability of $X$}
\label{sec:clusterability}


We start with a discussion using the smoothness and clusterability of $X$. Recent results have explored the clusterability of $X$s \cite{zhu2021clusterability,bahri2020deep} to infer the noise transition matrix using the clusterability of $X$:
\begin{definition}
The 2-NN clusterability requires each $X$ and its two nearest neighbors $X_1, X_2$ share the same true label $Y$, that is $Y=Y_1=Y_2$. \label{def:2nn}
\end{definition}

\begin{wrapfigure}{R}{0.4\textwidth}
\vspace{-5mm}
\begin{minipage}{0.4\textwidth}
    \centering
    \includegraphics[width=\textwidth]{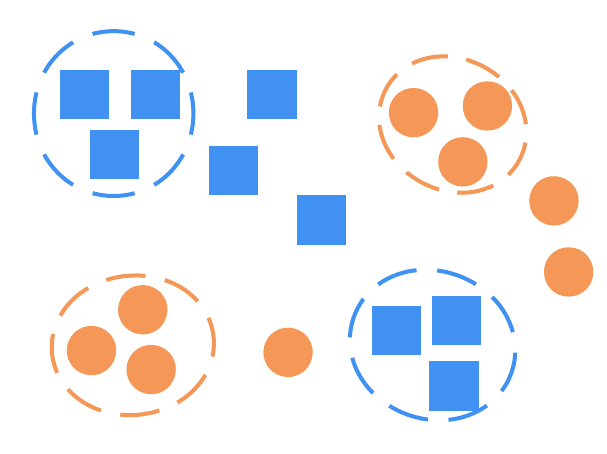}
    \vspace{-6mm}
    \caption{Data generation: label correlation among triplets.}
    \label{fig:causal_graph}
\end{minipage}
\vspace{-3mm}
\end{wrapfigure}

This definition effectively helps us avoid the need for multiple noisy labels per each $X$: one can view it as for each $X$, borrowing the noisy labels from its 2-NN, all together we have three independent noisy labels $\nY, \nY_1, \nY_2$, all given the same $Y$. This smoothness or clusterability condition allows us to apply our identifiability results when one believes the $T(X)$ stays the same for the 2-NN nearest neighborhood $X,X_1,X_2$.

But, when does an instance $X$ and its 2-NN $X_1,X_2$ share the same true label? This requirement seems strange at the first sight: as long as $\p(Y|X), \p(Y_1|X_1)$ are not degenerate (being either 0 or 1 for different label classes), there always seems to be a positive probability that the realized $Y \neq Y_1$, no matter how close $X$ and $X_1$ are. Nonetheless, empirically, the 2-NN requirement seems to hold well: according to \cite{zhu2021clusterability} (Table 3 therein), when using a feature extractor built using the clean label, more than 99\% of the instance satisfies the 2-NN condition. Even when using a feature extractor trained on noisy labels, the ratio is mostly always in or close to the $80\%$ range. 
 
The following data generation process for an unstructured discrete domain of classification problems \cite{feldman2020does,liu2021importance} helps us justify the 2-NN requirement. The intuition is that when $X$s are informative and sufficiently discriminative, the similar $X$s are going to enjoy the same true label.

\squishlist
    \item Let $\lambda = \{\lambda_1,...,\lambda_n\}$ denote the priors for each $X \in \mathcal X$.
    \item For each $X \in \mathcal X$, sample a quantity $q_X$ independently and uniformly from the set $\lambda$.
    \item The resulting probability mass function of $X$ is given by $D(X)=\frac{q_X}{\sum_{X \in \mathcal X} q_X}$. 
    \item A total of $N$ $X$s are observed. Denote by $X_1,X_2$  $X$'s two nearest neighbors. 
    \item Each $(X,X_1,X_2)$ forms a triplet if $||X_1-X||, ||X_2-X||$ fall below a threshold $\epsilon$ (closeness).
    \item A single $Y$ for the tuple $(X,X_1,X_2)$ draws from $\p(Y|X, X_1, X_2)$.
    \item Based on $Y$, we further observe three $\nY, \nY_1, \nY_2$ according to $\p(\nY, \nY_1, \nY_2|Y)$.
\squishend

The above data-generation process captures the correlation among $X$s that are really close. We prove the above data generation process satisfies the 2-NN clusterability requirement with high probability. 
\begin{theorem}
When $N$ is large enough such that $N> \frac{4\sum_{X \in \mathcal X} q_X}{\min_X q_X}$, w.p. at least $1 - N exp(-2 N)$, each $X$ and its two nearest neighbor $X_1, X_2$ satisfy the 2-NN criterion. \label{thm:2nn}
\end{theorem}

%
\paragraph{Smoothness conditions in semi-supervised learning} 
~This above discussion also ties closely to the smoothness requirements in semi-supervised learning \cite{zhu2003semi,zhu2005semi}, where the neighborhood $X$s can provide and propagate label information in each local neighborhood of $X$s. Indeed, this idea echoes the co-teaching solution \cite{jiang2018mentornet,han2018co} in the literature of learning with noisy labels, where a teacher/mentor network is trained to provide artificially generated noisy labels to supervise the training of the student network. Our identifiability result, to a certain degree, implies that the addition of the additional noisy supervision improves the chance for identifying $T(X)$.  
In \cite{jiang2018mentornet,han2018co}, counting the noisy label itself, and the ``teacher" supervision, there are two such noisy supervision labels. This observation raises an interesting question: since our result emphasized three labels, does adding an additional teacher network for an additional supervision help? This question merits empirical verification. One caution we want to make is with more artificially inserted noisy labels, likely a subset of them will start to become dependent on each other, due to the similarities in training different models. This points out yet another interesting identifiability problem whose solution will help us detect the dependent labels and introduce additional and appropriate variables for modeling these dependencies.


\subsection{Leveraging smoothness and clusterability of $T(X)$}
\label{sec:group1}

\begin{wrapfigure}{R}{0.35\textwidth}
\vspace{-5mm}
\begin{minipage}{0.35\textwidth}
    \centering
    \includegraphics[width=\textwidth]{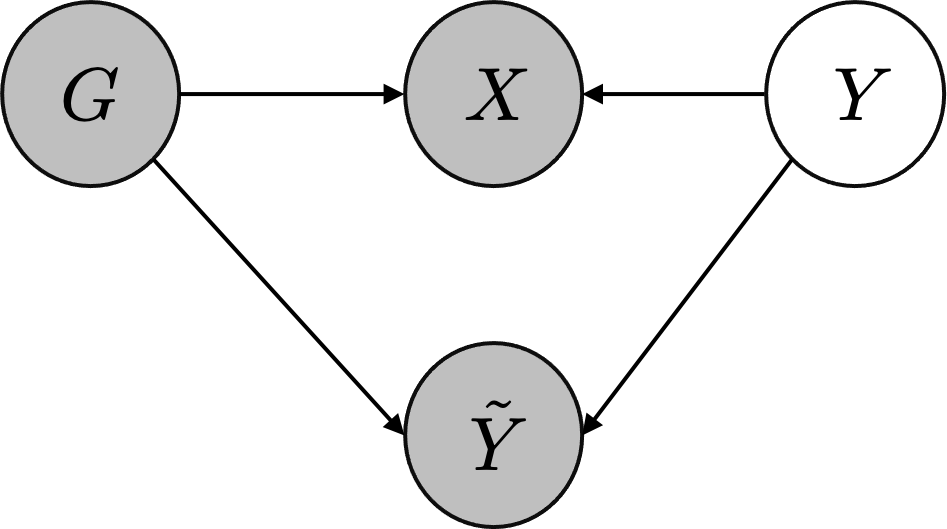}
    \vspace{-6mm}
    \caption{Causal graph for $(G, X, Y, \tilde{Y})$. Grey color indicates the variables are observable.}
    \label{fig:causal:knownG}
\end{minipage}
\vspace{-3mm}
\end{wrapfigure}

In this section, we show that another ``smoothness" assumption of $T(X)$ introduces new observation variables for us to identify $T(X)$. In Figure \ref{fig:causal:knownG}, we define variable $G=\{1,2,...,|G|\}$ to denote the group membership for each $X$. Consider a scenario that $X$ can be grouped into $|G|$ groups such that each group of $X$s share the same $T(X)$: $T(X_1) = T(X_2)$ if $X_1, X_2$ share the same group membership. We observe $G, X, \nY$. This type of grouping has been observed in the literature:

\noindent \textbf{Class-dependent $T$}
Clearly, when $T(X) \equiv T$, the entire set of $X$ can then be viewed as the group with $|G| = 1$.


\noindent \textbf{Noise clusterability}
The noise transition estimator proposed in \cite{zhu2021clusterability} was primarily developed for class-dependent but not instance-dependent $T(X)$. Nonetheless,  a noise clusterability definition is introduced in \cite{zhu2021clusterability} to allow the approach to be applied to the instance-dependent noise setting. 
Under the noise clusterability, using off-the-shelf clustering algorithms can help separate the dataset into local ones. 

\noindent \textbf{Group-dependent $T(X)$}
Recent results have also studied the case that the data $X$ can be grouped using additional information \cite{wang2021fair,liu2021can,wang2021learning}. For instance, \cite{wang2021fair,liu2021can} consider the setting where the data can be grouped by the associated ``sensitive information", e.g., by age, gender, or race. Then the noise transition matrix remains the same for $X$s that come from each group.

By this grouping, $X$ becomes informative observations for each hidden $Y$ and will fulfill the requirement of observing additional noisy labels. Below we formalize this discussion. We now define a disentangled feature and an informative feature: Denote by $R(X) \in \mathbb R^{d^*}$ a learned representation for $X$. Denote by $R_i$ the random variable for $R_i(X), i = 1,2,..,d^*$. For simplicity of the analysis, we assume each $R_i$ has finite observation space $\mathcal R_i$ with cardinality $|\mathcal R_i| = \kappa_i$. We believe the result has implications for continuous feature space but the formalization of the results will be left for future technical developments.  Define $M_i$ for each $R_i$ as
$
M_i[j,k] = \p(R_i = \mathcal R_i[k]|Y=j),
$
where in above $\mathcal R_i[k]$ denotes the $k$-th element in $\mathcal R_i$.
\begin{definition}[Disentangled $R$] $R$ is disentangled if $\{R_i\}_{i=1}^{d^*}$ are conditional independent given $Y$.
\end{definition}

\begin{definition}[Informative features] $R_i$ is informative if its Kruskal rank is at least 2: $\emph{\kr}(M_i) \geq 2$.
\end{definition}


Assuming each $X$ can be transformed into a set of disentangled features $R$, we prove:
\begin{theorem}\label{thm:group}
For $X$s in a given group $g \in G$, with a single informative noisy label, $T(X)$ is identifiable 
if the number of disentangled and informative features $d^*$ satisfy that $d^* \geq K$.
\end{theorem}

 This result points out a new observation that even when we have a single noisy label, given a sufficient number of disentangled and informative features, the noise transition matrix $T$ is indeed identifiable, without requiring either multiple noisy labels, or the anchor point condition.

The above result aligns with recent discussions of a neural network being able to disentangle features \cite{higgins2018towards,steenbrugge2018improving} proves to be a helpful property. We establish that having disentangled feature helps identify $T(X)$.
The required number of disentangled features grows linearly in $K$. When relaxing the unique identifiability to generic identifiability, i.e., the identifiability scenario has measure zero \cite{allman2009identifiability}, the above theorem can be further extended to requiring $d^* \geq \lceil \log_2 \frac{K+2}{2} \rceil$. We leave the details in Appendix (Theorem \ref{thm:generic:iden}). The above result is particularly informative when the number of classes $K$ is large.

When the disentangled feature is not given, how do we disentangle $X$ using only noisy labels to benefit from our results? In Section \ref{sec:exp} we will test the effectiveness of a self-supervised representation learning approach that takes the side information relative to true label $Y$ but operates independently from noisy labels. This result also implies when the noise rate is high such that $\nY$ starts to become uninformative, dropping the noisy labels and focusing on obtaining the disentangled features helps with the identifiability of $T(X)$. This observation also helps explain recent successes in applying semi-supervised \cite{cheng2020learning,Li2020DivideMix,nguyen2019self} and self-supervised \cite{cheng2021demystifying,zheltonozhskii2022contrast,ghosh2021contrastive} learning techniques to the problem of learning from noisy labels.




\subsection{Smoothness and clusterability of $T(X)$ with unknown groupings}
\label{sec:group2}

\begin{wrapfigure}{R}{0.35\textwidth}
\vspace{-5mm}
\begin{minipage}{0.35\textwidth}
    \centering
    \includegraphics[width=\textwidth]{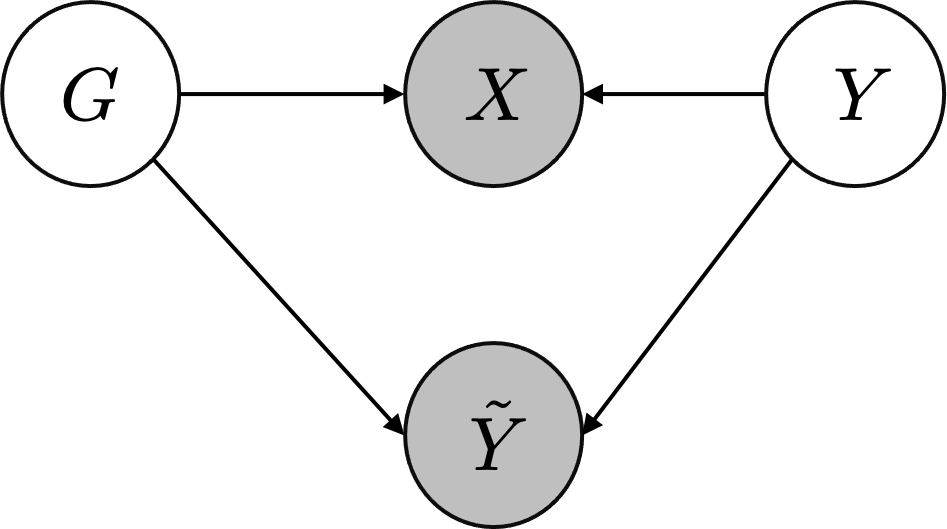}
    \vspace{-6mm}
    \caption{Causal graph with unobserved $G$. Grey indicates observable variables.}
    \label{fig:causal:unknownG}
\end{minipage}
\vspace{-3mm}
\end{wrapfigure}

In practice, we often do not know the groupings of $X$ that share the same $T(X)$, nor do we have a clear power (e.g., the noise clusterability condition) to separate the data into different groups. In reality, different from Figure \ref{fig:causal:knownG}, the group membership can often remain hidden, if no additional knowledge of the data is solicited, leading to a situation in Figure \ref{fig:causal:unknownG}.

It is a non-trivial task to jointly infer the group membership  with $T(X)$. We first show that mixing the group membership can lead to non-negligible estimation errors. Suppose that there are two groups of $X$, each having a noise transition matrix $T_1(X), T_2(X)$. Suppose we ended up estimating one $T^*(X)$ for both groups mistakenly. We then have:
\begin{theorem}\label{thm:est:error}
Any estimator $T^*(X)$ will incur at least the following estimation error: 
\vspace{-0.1in}
\begin{align}
||T_1(X)-T^*(X)||_F + ||T_2(X)-T^*(X)||_F  \geq \frac{1}{\sqrt{2}}||T_1(X)-T_2(X)||_F
\end{align}
\end{theorem}
\vspace{-0.1in}

The above result shows the necessity of identifying $G$ as well. Now we present our positive result on the identifiability when $G$ is hidden too: Re-number the combined space of $G \times Y$ as $\{1,2,...,|G| K\}$.
We are going to reuse the definition of $M_i$ for each disentangled feature $R_i$:  Define the ``Kruskal matrix" for each $R_i$ as
$
M_i[j,k] = \p(R_i = \mathcal R_i[k]|G \times Y=j).
$ 
\begin{theorem}
For $X$s in a given group $g \in G$, with a single informative noisy label, $T(X)$ is identifiable if the number of disentangled and informative features $d^*$ satisfy that $d^* \geq 2|G|K-1$.

\label{thm:iden:unknown:group}
\end{theorem}

When we have unknown groups of noise, the requirement of the number of informative and disentangled features grows linearly in $|G|$. We now relate to the literature that implicitly groups $X$s. We will use $\mathcal X$ to denote the space of all possible $X$s.

\noindent \textbf{Part-dependent label noise} 
\cite{xia2020parts} discusses a part-dependent label noise model where each $T(X)$ can decompose into a linear combination of $p$ parts:
$
    T(X) = \sum_{i}^{p} \omega_i(X) \cdot T_i.
$
The motivation of the above model is each $X$ can be viewed as a combination of multiple different sub-parts, and each of them has a certain difficulty being labeled.
The hope is that the parameter space $\omega(X)$ can reduce the dependency between $X$ and $T(X)$. Denote
$
    \mathcal W:=\{\omega(X): X \in \mathcal X\}.
$ To put into our result, $|G| = |\mathcal W|$. If $\mathcal W$ has a much smaller space than $\mathcal X$, the condition specified in Theorem \ref{thm:iden:unknown:group} would be more likely to be satisfied.

\noindent \textbf{DNN approach} \cite{yang2021estimating} proposes using a deep neural network to encode the dependency between $X$ and $T^*(X)$, with the only difference being that $T^*(X)$ is defined as the transition between $\nY$ and the Bayes optimal label $Y^*$. Define:
$
    \dnn:=\{\dnn(X): X \in \mathcal X\}.
$
Similarly, in analogy to our results in Theorem \ref{thm:iden:unknown:group}, with replacing the hidden variable $Y$ to $Y^*$, $|G|$ will be determined by $|\dnn|$. So long as the $\dnn$ can identify the patterns in $T(X)$ and compress the space of $\dnn(X)$ as compared to $\mathcal X$, the identifiability becomes easier to achieve.

\noindent \textbf{The causal approach} \cite{yao2021instance} proposed improving the identifiability by exploring the causal structure. 
With causal inference approaches, one can identify a more representative and compressed $\tilde{X}$ for each $X$ such that $\p(\tilde{Y}|Y,X,\tilde{X}) = \p(\tilde{Y}|Y,\tilde{X})$. 
Denote
$
    \tilde{\mathcal X}:=\{\tilde{X}: \tilde{X} \rightarrow X \in \mathcal X\}.
$
To plug in our results, we have
$|G| = |\tilde{\mathcal X}|$. 

\section{Some Empirical Evidence: Disentangled Features}\label{sec:exp}

Most of our results above verified the empirical success of existing approaches and we refer the interested reader to the detailed experiments in the corresponding references. 
We now empirically show the possibility of learning disentangled features to help identify the noise transition matrix. We consider three types of encoders that are used to generate features. The first encoder is pre-trained by cross-entropy (CE) loss via a weakly supervised manner which is generally adopted in FW \cite{patrini2017making} and HOC \cite{zhu2021clusterability}. However, since the training data is noisy, it is hard to guarantee that features are disentangled - this is our baseline. The second encoder is pre-trained by SimCLR \cite{chen2020simple} via a self-supervised manner. It is shown that the features trained by SimCLR are partly disentangled on some simple augmentation features such as rotation and colorization \cite{wang2021self}. The third encoder is trained by IPIRM \cite{wang2021self} via a self-supervised manner which can generate fully disentangled features. After training these three encoders, we fix the encoder and generate features from raw samples to estimate the noise transition matrix using HOC estimator \cite{zhu2021clusterability}. We evaluate the performance via absolute estimation error defined below:
$
    \text{err} = \frac{\sum_{i=1}^{K}\sum_{j=1}^{K}|\hat{T}_{i,j} - T_{i,j}|}{K^{2}} * 100,
$
where $\hat{T}$ is the estimated noise transition matrix, $T$ is the real noise-transition matrix, $K$ is the number of classes in the dataset, which is also the size of the transition matrix. The overall experiments are shown in Table \ref{hoc_estimator}. We can observe that the estimation error decreases as features become more disentangled which supports our analyses in the paper. We defer the details, more experiments, as well as experiments on comparing training performances using disentangled features, to the supplementary material.

\vspace{-0.1in}
\begin{scriptsize}
{ \begin{table*}[!h]
		\caption{Comparison of transition matrix estimation error for different types of features on CIFAR-10. Each experiment is run 3 times and mean $\pm$ std is reported. \emph{asymm.}: asymmetric label noise; \emph{inst.}: instance-dependent label noise. Numbers are noise rates.
		All the encoders are from ResNet50 backbone. }
		\begin{center}
			\begin{tabular}{c|ccccc} 
				\hline 
			  Feature Type & \emph{ asymm. 0.3} & \emph{ asymm. 0.4} & \emph{ inst. 0.4}
			  & \emph{ inst. 0.5}& \emph{ inst. 0.6}
			 \\ 
			 \hline\hline
				Weakly-Supervised &14.51 $\pm$ 0.4 & 15.2 $\pm$ 0.02 &8.39 $\pm$ 0.05 &6.91 $\pm$ 0.06 &6.18 $\pm$ 0.15\\
			     SimCLR &4.42 $\pm$ 0.01 &4.41 $\pm$ 0.01  & 2.91 $\pm$ 0.02&2.55$\pm$ 0.04 &2.64 $\pm$ 0.03\\
			     IPIRM & \textbf{3.73 $\pm$ 0.02}& \textbf{3.74 $\pm$ 0.01} & \textbf{2.47 $\pm$ 0.03} & \textbf{2.20 $\pm$ 0.02} & \textbf{2.37 $\pm$ 0.06}\\
			     \hline
			\end{tabular}
		\end{center}
		\label{hoc_estimator}
	\end{table*}
}

\end{scriptsize}

\vspace{-0.1in}


\section{Concluding Remarks}\label{sec:conclude}
This paper characterizes the identifiability of instance-level label noise transition matrix. We connect the problem to the celebrated Kruskal's identifiability result and present a necessary and sufficient condition for the instance-level identifiability. We extend and instantiate our results to practical settings to explain the successes of existing solutions. As a by-product, we show the importance of disentangled and informative features for identifying the noise transition matrix.

Future direction of work includes exploring the extension of our results to other weakly supervised learning settings (e.g., Positive Unlabeled learning, semi-supervised learning etc). Our results also encourage discussions on what assumptions are needed for the data in order to improve the identifiability of hidden factors.

\small
\bibliographystyle{unsrt}
\bibliography{ref}	

\begin{thebibliography}{10}

\bibitem{natarajan2013learning}
Nagarajan Natarajan, Inderjit~S Dhillon, Pradeep~K Ravikumar, and Ambuj Tewari.
\newblock Learning with noisy labels.
\newblock In {\em Advances in neural information processing systems}, pages
  1196--1204, 2013.

\bibitem{Patrini_2017_CVPR}
Giorgio Patrini, Alessandro Rozza, Aditya Krishna~Menon, Richard Nock, and
  Lizhen Qu.
\newblock Making deep neural networks robust to label noise: A loss correction
  approach.
\newblock In {\em The IEEE Conference on Computer Vision and Pattern
  Recognition (CVPR)}, July 2017.

\bibitem{wang2021fair}
Jialu Wang, Yang Liu, and Caleb Levy.
\newblock Fair classification with group-dependent label noise.
\newblock In {\em Proceedings of the 2021 ACM Conference on Fairness,
  Accountability, and Transparency}, FAccT '21, page 526–536, New York, NY,
  USA, 2021. Association for Computing Machinery.

\bibitem{han2020sigua}
Bo~Han, Gang Niu, Xingrui Yu, Quanming Yao, Miao Xu, Ivor Tsang, and Masashi
  Sugiyama.
\newblock Sigua: Forgetting may make learning with noisy labels more robust.
\newblock In {\em International Conference on Machine Learning}, pages
  4006--4016. PMLR, 2020.

\bibitem{zhu2021good}
Zhaowei Zhu, Zihao Dong, Hao Cheng, and Yang Liu.
\newblock A good representation detects noisy labels.
\newblock {\em arXiv preprint arXiv:2110.06283}, 2021.

\bibitem{liu2021can}
Yang Liu and Jialu Wang.
\newblock Can less be more? when increasing-to-balancing label noise rates
  considered beneficial.
\newblock NeurIPS'21.

\bibitem{xia2019anchor}
Xiaobo Xia, Tongliang Liu, Nannan Wang, Bo~Han, Chen Gong, Gang Niu, and
  Masashi Sugiyama.
\newblock Are anchor points really indispensable in label-noise learning?
\newblock {\em Advances in Neural Information Processing Systems}, 32, 2019.

\bibitem{zhu2021clusterability}
Zhaowei Zhu, Yiwen Song, and Yang Liu.
\newblock Clusterability as an alternative to anchor points when learning with
  noisy labels.
\newblock {\em ICML}, 2021.

\bibitem{scott2015rate}
Clayton Scott.
\newblock A rate of convergence for mixture proportion estimation, with
  application to learning from noisy labels.
\newblock In {\em AISTATS}, 2015.

\bibitem{liu2016classification}
Tongliang Liu and Dacheng Tao.
\newblock Classification with noisy labels by importance reweighting.
\newblock {\em IEEE Transactions on pattern analysis and machine intelligence},
  38(3):447--461, 2016.

\bibitem{yao2020dual}
Yu~Yao, Tongliang Liu, Bo~Han, Mingming Gong, Jiankang Deng, Gang Niu, and
  Masashi Sugiyama.
\newblock Dual t: Reducing estimation error for transition matrix in
  label-noise learning.
\newblock {\em arXiv preprint arXiv:2006.07805}, 2020.

\bibitem{cheng2017learning}
Jiacheng Cheng, Tongliang Liu, Kotagiri Ramamohanarao, and Dacheng Tao.
\newblock Learning with bounded instance-and label-dependent label noise.
\newblock In {\em Proceedings of the 37th International Conference on Machine
  Learning}, ICML '20, 2020.

\bibitem{northcutt2017learning}
Curtis~G Northcutt, Tailin Wu, and Isaac~L Chuang.
\newblock Learning with confident examples: Rank pruning for robust
  classification with noisy labels.
\newblock {\em UAI}, 2017.

\bibitem{northcutt2021confident}
Curtis Northcutt, Lu~Jiang, and Isaac Chuang.
\newblock Confident learning: Estimating uncertainty in dataset labels.
\newblock {\em Journal of Artificial Intelligence Research}, 70:1373--1411,
  2021.

\bibitem{liu2018surrogate}
Yang Liu, Juntao Wang, and Yiling Chen.
\newblock Surrogate scoring rules and a dominant truth serum.
\newblock {\em ACM EC}, 2020.

\bibitem{zhang2021learning}
Yivan Zhang, Gang Niu, and Masashi Sugiyama.
\newblock Learning noise transition matrix from only noisy labels via total
  variation regularization.
\newblock {\em arXiv preprint arXiv:2102.02414}, 2021.

\bibitem{li2021provably}
Xuefeng Li, Tongliang Liu, Bo~Han, Gang Niu, and Masashi Sugiyama.
\newblock Provably end-to-end label-noise learning without anchor points.
\newblock {\em arXiv preprint arXiv:2102.02400}, 2021.

\bibitem{xia2020parts}
Xiaobo Xia, Tongliang Liu, Bo~Han, Nannan Wang, Mingming Gong, Haifeng Liu,
  Gang Niu, Dacheng Tao, and Masashi Sugiyama.
\newblock Part-dependent label noise: Towards instance-dependent label noise.
\newblock {\em Advances in Neural Information Processing Systems},
  33:7597--7610, 2020.

\bibitem{cheng2020learning}
Hao Cheng, Zhaowei Zhu, Xingyu Li, Yifei Gong, Xing Sun, and Yang Liu.
\newblock Learning with instance-dependent label noise: A sample sieve
  approach.
\newblock In {\em International Conference on Learning Representations}, 2021.

\bibitem{yao2021instance}
Yu~Yao, Tongliang Liu, Mingming Gong, Bo~Han, Gang Niu, and Kun Zhang.
\newblock Instance-dependent label-noise learning under a structural causal
  model.
\newblock {\em Advances in Neural Information Processing Systems}, 34, 2021.

\bibitem{kruskal1976more}
Joseph~B Kruskal.
\newblock More factors than subjects, tests and treatments: an indeterminacy
  theorem for canonical decomposition and individual differences scaling.
\newblock {\em Psychometrika}, 41(3):281--293, 1976.

\bibitem{kruskal1977three}
Joseph~B Kruskal.
\newblock Three-way arrays: rank and uniqueness of trilinear decompositions,
  with application to arithmetic complexity and statistics.
\newblock {\em Linear algebra and its applications}, 18(2):95--138, 1977.

\bibitem{patrini2017making}
Giorgio Patrini, Alessandro Rozza, Aditya Krishna~Menon, Richard Nock, and
  Lizhen Qu.
\newblock Making deep neural networks robust to label noise: A loss correction
  approach.
\newblock In {\em Proceedings of the IEEE Conference on Computer Vision and
  Pattern Recognition}, pages 1944--1952, 2017.

\bibitem{liu2015classification}
Tongliang Liu and Dacheng Tao.
\newblock Classification with noisy labels by importance reweighting.
\newblock {\em IEEE Transactions on pattern analysis and machine intelligence},
  38(3):447--461, 2015.

\bibitem{xiao2015learning}
Tong Xiao, Tian Xia, Yi~Yang, Chang Huang, and Xiaogang Wang.
\newblock Learning from massive noisy labeled data for image classification.
\newblock In {\em Proceedings of the IEEE Conference on Computer Vision and
  Pattern Recognition}, pages 2691--2699, 2015.

\bibitem{berthon2020confidence}
Antonin Berthon, Bo~Han, Gang Niu, Tongliang Liu, and Masashi Sugiyama.
\newblock Confidence scores make instance-dependent label-noise learning
  possible.
\newblock {\em arXiv preprint arXiv:2001.03772}, 2020.

\bibitem{yang2021estimating}
Shuo Yang, Erkun Yang, Bo~Han, Yang Liu, Min Xu, Gang Niu, and Tongliang Liu.
\newblock Estimating instance-dependent label-noise transition matrix using
  dnns.
\newblock {\em arXiv preprint arXiv:2105.13001}, 2021.

\bibitem{zhu2020second}
Zhaowei Zhu, Tongliang Liu, and Yang Liu.
\newblock A second-order approach to learning with instance-dependent label
  noise.
\newblock {\em CVPR}, 2021.

\bibitem{berthon2021confidence}
Antonin Berthon, Bo~Han, Gang Niu, Tongliang Liu, and Masashi Sugiyama.
\newblock Confidence scores make instance-dependent label-noise learning
  possible.
\newblock In {\em International Conference on Machine Learning}, pages
  825--836. PMLR, 2021.

\bibitem{xia2020robust}
Xiaobo Xia, Tongliang Liu, Bo~Han, Chen Gong, Nannan Wang, Zongyuan Ge, and
  Yi~Chang.
\newblock Robust early-learning: Hindering the memorization of noisy labels.
\newblock In {\em International Conference on Learning Representations}, 2020.

\bibitem{Li2020DivideMix}
Junnan Li, Richard Socher, and Steven~C.H. Hoi.
\newblock Dividemix: Learning with noisy labels as semi-supervised learning.
\newblock In {\em International Conference on Learning Representations}, 2020.

\bibitem{yao2020towards}
Yu~Yao, Tongliang Liu, Bo~Han, Mingming Gong, Gang Niu, Masashi Sugiyama, and
  Dacheng Tao.
\newblock Towards mixture proportion estimation without irreducibility.
\newblock {\em arXiv preprint arXiv:2002.03673}, 2020.

\bibitem{menon2015learning}
Aditya Menon, Brendan Van~Rooyen, Cheng~Soon Ong, and Bob Williamson.
\newblock Learning from corrupted binary labels via class-probability
  estimation.
\newblock In {\em International Conference on Machine Learning}, pages
  125--134, 2015.

\bibitem{sidiropoulos2000uniqueness}
Nicholas~D Sidiropoulos and Rasmus Bro.
\newblock On the uniqueness of multilinear decomposition of n-way arrays.
\newblock {\em Journal of chemometrics}, 14(3):229--239, 2000.

\bibitem{allman2009identifiability}
Elizabeth~S Allman, Catherine Matias, and John~A Rhodes.
\newblock Identifiability of parameters in latent structure models with many
  observed variables.
\newblock {\em The Annals of Statistics}, 37(6A):3099--3132, 2009.

\bibitem{blanchard2010semi}
Gilles Blanchard, Gyemin Lee, and Clayton Scott.
\newblock Semi-supervised novelty detection.
\newblock {\em The Journal of Machine Learning Research}, 11:2973--3009, 2010.

\bibitem{clogg1995latent}
Clifford~C Clogg.
\newblock Latent class models.
\newblock In {\em Handbook of statistical modeling for the social and
  behavioral sciences}, pages 311--359. Springer, 1995.

\bibitem{anandkumar2014tensor}
Animashree Anandkumar, Rong Ge, Daniel Hsu, Sham~M Kakade, and Matus Telgarsky.
\newblock Tensor decompositions for learning latent variable models.
\newblock {\em Journal of machine learning research}, 15:2773--2832, 2014.

\bibitem{liu2012variational}
Qiang Liu, Jian Peng, and Alexander~T Ihler.
\newblock Variational inference for crowdsourcing.
\newblock {\em Advances in neural information processing systems}, 25:692--700,
  2012.

\bibitem{shah:nips13}
David~R Karger, Sewoong Oh, and Devavrat Shah.
\newblock Iterative learning for reliable crowdsourcing systems.
\newblock In {\em Advances in neural information processing systems}, pages
  1953--1961, 2011.

\bibitem{sig15}
Yang Liu and Mingyan Liu.
\newblock An online learning approach to improving the quality of
  crowd-sourcing.
\newblock In {\em Proceedings of the 2015 ACM SIGMETRICS International
  Conference on Measurement and Modeling of Computer Systems}, SIGMETRICS '15,
  pages 217--230, New York, NY, USA, 2015. ACM.

\bibitem{bahri2020deep}
Dara Bahri, Heinrich Jiang, and Maya Gupta.
\newblock Deep k-nn for noisy labels.
\newblock In {\em International Conference on Machine Learning}, pages
  540--550. PMLR, 2020.

\bibitem{feldman2020does}
Vitaly Feldman.
\newblock Does learning require memorization? a short tale about a long tail.
\newblock In {\em Proceedings of the 52nd Annual ACM SIGACT Symposium on Theory
  of Computing}, pages 954--959, 2020.

\bibitem{liu2021importance}
Yang Liu.
\newblock Understanding instance-level label noise: Disparate impacts and
  treatments, 2021.

\bibitem{zhu2003semi}
Xiaojin Zhu, Zoubin Ghahramani, and John~D Lafferty.
\newblock Semi-supervised learning using gaussian fields and harmonic
  functions.
\newblock In {\em Proceedings of the 20th International conference on Machine
  learning (ICML-03)}, pages 912--919, 2003.

\bibitem{zhu2005semi}
Xiaojin Zhu.
\newblock {\em Semi-supervised learning with graphs}.
\newblock Carnegie Mellon University, 2005.

\bibitem{jiang2018mentornet}
Lu~Jiang, Zhengyuan Zhou, Thomas Leung, Li-Jia Li, and Li~Fei-Fei.
\newblock Mentornet: Learning data-driven curriculum for very deep neural
  networks on corrupted labels.
\newblock In {\em International Conference on Machine Learning}, pages
  2304--2313. PMLR, 2018.

\bibitem{han2018co}
Bo~Han, Quanming Yao, Xingrui Yu, Gang Niu, Miao Xu, Weihua Hu, Ivor Tsang, and
  Masashi Sugiyama.
\newblock Co-teaching: Robust training of deep neural networks with extremely
  noisy labels.
\newblock In {\em Advances in neural information processing systems}, pages
  8527--8537, 2018.

\bibitem{wang2021learning}
Qizhou Wang, Jiangchao Yao, Chen Gong, Tongliang Liu, Mingming Gong, Hongxia
  Yang, and Bo~Han.
\newblock Learning with group noise.
\newblock {\em arXiv preprint arXiv:2103.09468}, 2021.

\bibitem{higgins2018towards}
Irina Higgins, David Amos, David Pfau, Sebastien Racaniere, Loic Matthey,
  Danilo Rezende, and Alexander Lerchner.
\newblock Towards a definition of disentangled representations.
\newblock {\em arXiv preprint arXiv:1812.02230}, 2018.

\bibitem{steenbrugge2018improving}
Xander Steenbrugge, Sam Leroux, Tim Verbelen, and Bart Dhoedt.
\newblock Improving generalization for abstract reasoning tasks using
  disentangled feature representations.
\newblock {\em arXiv preprint arXiv:1811.04784}, 2018.

\bibitem{nguyen2019self}
Duc~Tam Nguyen, Chaithanya~Kumar Mummadi, Thi Phuong~Nhung Ngo, Thi Hoai~Phuong
  Nguyen, Laura Beggel, and Thomas Brox.
\newblock Self: Learning to filter noisy labels with self-ensembling.
\newblock {\em arXiv preprint arXiv:1910.01842}, 2019.

\bibitem{cheng2021demystifying}
Hao Cheng, Zhaowei Zhu, Xing Sun, and Yang Liu.
\newblock Demystifying how self-supervised features improve training from noisy
  labels.
\newblock {\em arXiv preprint arXiv:2110.09022}, 2021.

\bibitem{zheltonozhskii2022contrast}
Evgenii Zheltonozhskii, Chaim Baskin, Avi Mendelson, Alex~M Bronstein, and
  Or~Litany.
\newblock Contrast to divide: Self-supervised pre-training for learning with
  noisy labels.
\newblock In {\em Proceedings of the IEEE/CVF Winter Conference on Applications
  of Computer Vision}, pages 1657--1667, 2022.

\bibitem{ghosh2021contrastive}
Aritra Ghosh and Andrew Lan.
\newblock Contrastive learning improves model robustness under label noise.
\newblock In {\em Proceedings of the IEEE/CVF Conference on Computer Vision and
  Pattern Recognition}, pages 2703--2708, 2021.

\bibitem{chen2020simple}
Ting Chen, Simon Kornblith, Mohammad Norouzi, and Geoffrey Hinton.
\newblock A simple framework for contrastive learning of visual
  representations.
\newblock In {\em International conference on machine learning}, pages
  1597--1607. PMLR, 2020.

\bibitem{wang2021self}
Tan Wang, Zhongqi Yue, Jianqiang Huang, Qianru Sun, and Hanwang Zhang.
\newblock Self-supervised learning disentangled group representation as
  feature.
\newblock {\em Advances in Neural Information Processing Systems}, 34, 2021.

\bibitem{wei2020combating}
Hongxin Wei, Lei Feng, Xiangyu Chen, and Bo~An.
\newblock Combating noisy labels by agreement: A joint training method with
  co-regularization.
\newblock In {\em Proceedings of the IEEE/CVF Conference on Computer Vision and
  Pattern Recognition}, pages 13726--13735, 2020.

\bibitem{he2016deep}
Kaiming He, Xiangyu Zhang, Shaoqing Ren, and Jian Sun.
\newblock Deep residual learning for image recognition.
\newblock In {\em Proceedings of the IEEE conference on computer vision and
  pattern recognition}, pages 770--778, 2016.

\bibitem{van2018representation}
Aaron Van~den Oord, Yazhe Li, and Oriol Vinyals.
\newblock Representation learning with contrastive predictive coding.
\newblock {\em arXiv e-prints}, pages arXiv--1807, 2018.

\bibitem{arjovsky2019invariant}
Martin Arjovsky, L{\'e}on Bottou, Ishaan Gulrajani, and David Lopez-Paz.
\newblock Invariant risk minimization.
\newblock {\em arXiv preprint arXiv:1907.02893}, 2019.

\end{thebibliography}
\appendix
\newpage
	
\section*{{\Large Appendix:
Identifiability of Label Noise Transition Matrix}}

The Appendix is organized in the following way: Section \ref{appendix_sec1} proves the Theorems in the main paper; Section \ref{appendix_sec2} provides more discussions on generic identifiability; Section \ref{appendix_sec3} provides more experiments on learning with noisy labels \emph{w.r.t.} disentangled features and elaborates the detailed experimental settings in the paper.

\section{Omitted Proofs}\label{appendix_sec1}

\section*{Proof for Lemma \ref{lemma:mpe}}
\begin{proof}
Using Bayes rule we easily obtain 
\begin{align}
    \p(X|\tilde{Y}=+1) &=\p(X|Y=+1) \cdot \p(Y=+1|\tilde{Y}=+1) \nonumber \\
    &+ \p(X|Y=-1)\cdot \p(Y=-1|\tilde{Y}=+1) \label{eqn:mpe1} 
\end{align}
The equality is due to the fact that $\tilde{Y}$ and $X$ are assumed to be independent given $Y$.
 Similarly:
\begin{align}
            \p(X|\tilde{Y}=-1) &=\p(X|Y=+1)  \cdot\p(Y=+1|\tilde{Y}=-1) \nonumber \\
            &+ \p(X|Y=-1) \cdot \p(Y=-1|\tilde{Y}=-1) \label{eqn:mpe2} 
\end{align}
Denote by $\pi_+:=\p(Y=-1|\tilde{Y}=+1), \pi_- := \p(Y=+1|\tilde{Y}=-1)$.
Since both $\p(X|Y=+1), \p(X|Y=-1)$ are unknown, 
solving Eqn. (\ref{eqn:mpe1}) and (\ref{eqn:mpe2}) we further have 
\begin{align}
  \p(X|\tilde{Y}=-1) = \tilde{\pi}_- \cdot \p(X|\tilde{Y}=+1) + (1-\tilde{\pi}_-)\cdot\p(X|Y=-1)\\
    \p(X|\tilde{Y}=+1) = \tilde{\pi}_+ \cdot \p(X|\tilde{Y}=-1) + (1-\tilde{\pi}_+)\cdot\p(X|Y=+1).
\end{align}
\end{proof}

\section*{Proof for Theorem \ref{thm:equiv}}

\begin{proof}
Further from  $\tilde{\pi}_-, \tilde{\pi}_+$ we can solve and derive $\pi_- = \frac{\tilde{\pi}_- (1-\tilde{\pi}_+)}{1-\tilde{\pi}_- \tilde{\pi}_+}, \pi_+ = \frac{\tilde{\pi}_+ (1-\tilde{\pi}_-)}{1-\tilde{\pi}_- \tilde{\pi}_+}$, establishing the equivalence between identifying  $\tilde{\pi}_-, \tilde{\pi}_+$ with identifying $\pi_-, \pi_+$. Next we show that identifying $\pi_-, \pi_+$ is equivalent with identifying $\{e_+,e_-\}$.

We first show identifying $\{\pi_+, \pi_-\}$ suffices to identify $\{e_+,e_-\}$. To see this, 
\begin{align*}
    \p(\tilde{Y}=+1|Y=-1) = \frac{\p(Y=-1|\tilde{Y}=+1)\p(\tilde{Y}=+1)}{\p(Y=-1)}
\end{align*}
And:
\begin{align*}
    \p(Y=-1) &= \p(Y=-1|\tilde{Y}=+1)\p(\tilde{Y}=+1) + \p(Y=-1|\tilde{Y}=-1)\p(\tilde{Y}=-1)
\end{align*}
The derivation for $  \p(\tilde{Y}=-1|Y=+1)$ is entirely symmetric.
Since we directly observe $\p(\tilde{Y}=-1), \p(\tilde{Y}=+1)$, with identifying $\p(Y=+1|\tilde{Y}=-1),\p(Y=-1|\tilde{Y}=+1)$, we can identify $\p(\tilde{Y}=+1|Y=-1),\p(\tilde{Y}=-1|Y=+1)$. 

Next we show that  to identify $\{e_+,e_-\}$, it is necessary to identify $\{\pi_+, \pi_-\}$. Suppose not: we are unable to identify $\pi_+,\pi_i$ but are able to identify $\{e_+,e_-\}$. This implies that there exists another pair $\{\pi'_+, \pi'_-\} \neq \{\pi_+, \pi_-\}$ such that (denote by $\tilde{p}:= \p(\tilde{Y}=+1)$)
\begin{align}
    \p(\tilde{Y}=+1|Y=-1) &= \frac{\pi_+ \tilde{p} }{\pi_+ \tilde{p}+(1-\pi_-) (1-\tilde{p})} 
\label{eqn:e-:1}    \\
    &=\frac{\pi'_+ \tilde{p}}{\pi'_+ \tilde{p}+(1-\pi'_-) (1-\tilde{p})}\label{eqn:e-:2}
\end{align}
\begin{align}
    \p(\tilde{Y}=-1|Y=+1) &= \frac{\pi_- (1-\tilde{p}) }{(1-\pi_+) \tilde{p}+ \pi_- (1-\tilde{p})}\label{eqn:e+:1}\\
    &=\frac{\pi'_- (1-\tilde{p}) }{(1-\pi'_+) \tilde{p}+ \pi'_- (1-\tilde{p})}\label{eqn:e+:2}
\end{align}
By dividing $\pi_+, \pi'_+$ in both the numerator and denominator in Eqn. (\ref{eqn:e-:1}) and (\ref{eqn:e-:2}), we conclude that 
\begin{align}
    \frac{1-\pi_-}{\pi_+} = \frac{1-\pi'_-}{\pi'_+}\label{eqn:iden1}
\end{align}


While from Eqn. (\ref{eqn:e+:1}) and (\ref{eqn:e+:2}) we conclude

\begin{align}
    \frac{1-\pi_+}{\pi_-} = \frac{1-\pi'_+}{\pi'_-} \label{eqn:iden2}
\end{align}
From Eqn. (\ref{eqn:iden1}) and (\ref{eqn:iden2}) we have 
\begin{align}
    (1-\pi_-)\pi'_+ = (1-\pi'_-)\pi_+\\
    (1-\pi'_+)\pi_- = (1-\pi_+)\pi'_-
\end{align}
Taking the difference and re-arrange terms we prove
\[
\pi_+ + \pi_- = \pi'_+ + \pi'_- 
\]
From Eqn. (\ref{eqn:iden1}) again, taking $-1$ on both side we have
\begin{align}
        \frac{1-\pi_- - \pi_+}{\pi_+} = \frac{1-\pi'_--\pi'_+}{\pi'_+}
\end{align}
This proves $\pi_+ = \pi'_+$. Similarly we have $\pi_- = \pi'_-$ - but this contradicts the assumption that $\{\pi'_-, \pi'_+\}$ is a different pair. 
\end{proof}

\section*{Proof for Theorem \ref{thm:main:instance}}

\begin{proof} 
We first prove sufficiency. We first relate our problem setting to the setup of Kruskal's identifiability scenario: $Y \in \{1,2,...,K\}$ corresponds to the unobserved hidden variable $Z$. $\p(Y = i)$ corresponds to the prior of this hidden variable. Each $\nY_i, i=1,...,p$ corresponds to the observation $O_i$. $\kappa_i$ is then simply the cardinality of the noisy label space, $K$. In the context of this theorem, $p=3$, corresponding to the three noisy labels we have. 

Each $\nY_i$ corresponds to an observation matrix $M_i$:
\[
M_i[j,k] = \p(O_i = k|Z=j) = \p(\nY_i = k|Y=j, X)
\]
Therefore, by definition of $M_1,M_2,M_3$ and $T(X)$, they all equal to $T(X)$: $M_i \equiv T(X), i = 1,2,3$. When $T(X)$ has full rank, we know immediately that all rows in $M_1,M_2,M_3$ are independent. Therefore, the Kruskal ranks satisfy
\[
\kr(M_1) = \kr(M_2) = \kr(M_3) = K
\]
Checking the condition in Theorem \ref{thm:Kruskal}, we easily verify
\begin{align*}
  \kr(M_1) + \kr(M_2) + \kr(M_3) = 3K \geq 2K+2  
\end{align*}
Calling Theorem \ref{thm:Kruskal} proves the sufficiency. 

Now we prove necessity. To prove so, we are allowed to focus on the binary case, where 
\[
T(X) = \begin{bmatrix}
1-e_-(X) & e_-(X) \\
e_+(X) & 1-e_+(X)
\end{bmatrix}
\]
Note in above, for simplicity we drop $e_-,e_+$'s dependency in $X$.
We need to prove less than 3 informative labels will not suffice to guarantee identifiability. The idea is to show that the two different set  of parameters $e_-,e_+$ can lead to the same joint distribution $\p(\nY_1,\nY_2|X)$.

The case with a single label is already proved by Example 1. Now consider two noisy labels $\nY_1,\nY_2$. We first claim the following three quantities fully capture the information provided by $\nY_1,\nY_2$:
\squishlist
    \item Posterior: $\p(\nY_1 = +1|X)$ 
    \item Positive Consensus: $\p(\nY_1=\nY_2 = +1|X)$
    \item Negative Consensus: $\p(\nY_1=\nY_2 = -1|X)$
\squishend
This is because other statistics in $\nY_1,\nY_2|X$ can be reproduced using combinations of the three quantities above: 
\begin{align*}
&\p(\nY_1 = -1|X) = 1- \p(\nY_1 = +1|X)~,\\
&\p(\nY_1=+1, \nY_2 = -1|X) =  \p(\nY_1 = +1|X)-\p(\nY_1=\nY_2 = +1|X)~,\\
&\p(\nY_1=-1, \nY_2 = +1|X) =  \p(\nY_2 = +1|X) -\p(\nY_1=\nY_2 =+1|X)~.
\end{align*}
But $\p(\nY_2 = +1|X) = \p(\nY_1 = +1|X)$, since the two noisy labels are identically distributed. The above three quantities led to three equations that depend on $e_+, e_-$: denote by $\gamma:=\p(Y=+1)$

Next we prove the following system of equations:
\begin{align*}
    \p(\tilde{Y}=+1|X) &= \gamma \cdot (1-e_+) + (1-\gamma) \cdot e_-\\
    \p(\nY_1=\nY_2 = +1|X) &= \gamma \cdot (1-e_+)^2 + (1-\gamma) \cdot e^2_-\\
    \p(\nY_1=\nY_2 = -1|X) &= \gamma \cdot e_+^2 + (1-\gamma) \cdot (1- e_-)^2 
\end{align*}

To see this:
\begin{align*}
        &\p(\nY_1=\nY_2 = +1|X) \\
        =&\p(\nY_1=\nY_2 = +1, Y=+1|X)\\
        &+\p(\nY_1=\nY_2 = +1, Y=-1|X)\\
        =&\p(\nY_1=\nY_2 = +1| Y=+1, X)\cdot \p(Y=+1|X)\\
        &+\p(\nY_1=\nY_2 = +1| Y=-1, X) \cdot \p( Y=-1|X)\\
        =&\gamma\cdot (1-e_+)^2 + (1-\gamma) \cdot e^2_-
\end{align*}
The last equality uses the fact that $\nY_1, \nY_2$ are conditional independent given $Y$, so
\begin{align*}
    &\p(\nY_1=\nY_2 = +1| Y=+1, X) = \\
    &\p(\nY_1=+1| Y=+1, X)\cdot\p(\nY_2=+1| Y=+1, X)\\
    &\p(\nY_1=\nY_2 = +1| Y=-1, X)=\\
    &\p(\nY_1=+1| Y=-1, X)\cdot\p(\nY_2=+1| Y=-1, X)
\end{align*}
We can similarly derive for $\p(\nY_1=\nY_2 = -1|X)$.

Now we show the above equations do not identify $e_+,e_-$. For instance, it is straightforward to verify that both of the solutions below satisfy the equations (up to numerical errors, exact solution exists but in complicated forms):
\begin{itemize}
    \item $\gamma = 0.7, ~e_+ = 0.2, ~e_- = 0.2$
    \item $\gamma = 0.8,~ e_+ = 0.242, ~e_- = 0.07$
\end{itemize}
The above example proves that two informative noisy labels are insufficient to guarantee identifiability. 

\end{proof}


\section*{Proof for Theorem \ref{thm:2nn}}

\begin{proof}
In the unstructured model, we first show that, with a large $N$, with high probability, each $X$'s will present at least 3 times. Denote by $N_X$ the number of times $X$ appears in the dataset. Then
\begin{align}
    N_X := \sum_{i=1}^N 1[X_i = X], ~\E[ N_X] = \frac{q_X}{\sum_{X \in \mathcal X} q_X} N
\end{align}
When $N$ is large enough such that $N > \frac{4\sum_{X \in \mathcal X} q_X}{\min_X q_X}$, we have $\E[  N_X] > 4$.
Then using Hoeffding inequality we have 
\begin{align*}
 \p(N_X \leq 3) \leq exp(-2 N).
\end{align*}
Using union bound (across $N$ samples), it implies that with probability at least $1-N exp(-2 N)$, $N_X \geq 3, \forall X$:
\begin{align}
        \p(N_X > 3, \forall X) &= 1 - \p(N_X \leq 3, \exists X)\leq 1-N exp(-2 N)
\end{align}
This further implies that with probability at least $1-N exp(-2 N)$, we have $X_1=X_2=X$ for each $X$:
Their distance is 0, clearly falling below the closeness threshold $\epsilon$. Therefore they will share the same true label. 
\end{proof}

\section*{Proof for Theorem \ref{thm:group}}
\begin{proof}
The $d^*$ features and the noisy label $\nY$ jointly give us $d^*+1$ independent observations. Checking Kruskal's condition we have:
\begin{align*}
    & \kr(T(X)) + \sum_{i=1}^{d^*} \kr(M_i) \geq K + 2\cdot d^*\geq K + K + d^*= 2K + d^*+1-1
\end{align*}
Calling Theorem \ref{thm:Kruskal}, we establish the identifiability. 
\end{proof}

\section*{Proof for Theorem \ref{thm:est:error}}
\begin{proof}
By definition
\begin{align}
    ||T_1(X)-T^*(X)||_F = \sqrt{\sum_i \sum_j (T_1[i,j]-T[i,j])^2}
\end{align}
Easy to show that
\begin{align*}
    &||T_1(X)-T^*(X)||_F + ||T_2(X)-T^*(X)||_F\\
    =&\sqrt{\sum_i \sum_j (T_1[i,j]-T[i,j])^2} +\sqrt{\sum_i \sum_j (T_2[i,j]-T[i,j])^2}\\
    = &\sqrt{\left( \sqrt{\sum_i \sum_j (T_1[i,j]-T[i,j])^2} +\sqrt{\sum_i \sum_j (T_2[i,j]-T[i,j])^2}\right)^2}\\
        \geq &\sqrt{\sum_i \sum_j \left(\left (T_1[i,j]-T[i,j]\right)^2 + \left(T_2[i,j]-T[i,j]\right)^2 \right)} \tag{Dropping the cross-product term which is positive}\\
    \geq &\sqrt{\sum_i \sum_j \left (T_1[i,j]-\frac{T_1[i,j]+T_2[i,j]}{2}\right)^2 +  \left(T_2[i,j]-\frac{T_1[i,j]+T_2[i,j]}{2}\right)^2} \tag{minimum distance is at half}\\
    =&    \sqrt{\sum_i \sum_j 2 \left(\frac{T_1[i,j]-T_2[i,j]}{2}\right)^2}\\
    =&\frac{1}{\sqrt{2}} \sqrt{\sum_i \sum_j (T_1[i,j]-T_2[i,j])^2}\\
    =&\frac{1}{\sqrt{2}} ||T_1(X)-T_2(X)||_F
\end{align*}
\end{proof}

\section*{Proof for Theorem \ref{thm:iden:unknown:group}}
\begin{proof}
The proof is straightforward by checking KrusKal's identifiability condition:
\begin{align*}
    & \kr(T(X)) + \sum_{i=1}^{d^*} \kr(M_i) \geq 1  + 2\cdot d^* \geq 1  + 2|G| K-1 + d^* = 2|G| \cdot K + d^*+1 -1
\end{align*}
\end{proof}



\section{Generic identifiability}\label{appendix_sec2}
We provide a bit more detail for the discussion on generic identifiability left in Section \ref{sec:group1}.
\begin{theorem}
With a single informative noisy label, $T(X)$ is generically identifiable for each group $g \in G$ if the number of disentangled features $d^*$ satisfies that $d^* \geq \lceil \log_2 \frac{K+2}{2} \rceil$, and $\tau_i \geq 2$.
\label{thm:generic:iden}
\end{theorem}
\begin{proof}
We first reproduce a relevant theorem in \cite{allman2009identifiability}:
\begin{theorem}\cite{allman2009identifiability}
When $p=3$ (3 independently observations), the model parameters are generically identifiable, up to label permutation, if 
\begin{align}
    \min(K,\kappa_1) + \min(K,\kappa_2) + \min(K,\kappa_3) \geq 2K+2
\end{align}
\end{theorem}

Based on the above theorem we have the following identifiability result:

Grouping $d^*$ features evenly into two groups, each corresponding to a meta variable/feature:
\begin{align*}
    R^*_1 = \prod_{i=1}^{d^*_1} R_i, ~X^*_2 = \prod_{j=d^*_1+1}^{d^*} R_j
\end{align*}
Denote feature dimensions of each group as $d^*_1, d^*_2$:
\begin{align}
    \tau^*_1 = \prod_{i=1}^{d^*_1} \geq 2^{d^*_1} \geq 2^{\lceil \log_2 \frac{K+2}{2} \rceil} \geq \frac{K+2}{2}
\end{align}

Similarly $\tau^*_2 \geq \frac{K+2}{2}$. Denote by $M^*_1, M^*_2$ the two observation matrices for the grouped variables
\begin{align*}
M^*_i[j,k] = \p(R^*_i = \mathcal R^*_i[k]|Y=j), ~i=1,2.
\end{align*}
Then: 
\begin{align*}
    \kr(T(X)) + \kr(M^*_1) + \kr(M^*_2) \geq K + 2 \frac{K+2}{2} = 2K+2,
\end{align*}
which again satisfied the identifiability condition specified in Theorem \ref{thm:Kruskal}.
\end{proof}

\section{More experiments}\label{appendix_sec3}
In this section, we elaborate the detailed experiment setting and perform more experiments \emph{w.r.t.} disentangled features.

\subsection{Experiment setting for Table \ref{hoc_estimator}}

\textbf{Label Noise Generation} The label noise of each instance is characterized by $T_{ij}(X) = \mathbb P(\widetilde{Y}=j|X,Y=i)$. 
In this paper, we consider two types of label noise: asymmetric label noise \cite{han2018co,wei2020combating} and instance-dependent label noise \cite{cheng2020learning,zhu2020second}. For asymmetric label noise, $T(X)\equiv T$, 
each clean label is randomly flipped to its adjacent label w.p. $\epsilon$, where $\epsilon$ is the noise rate, i.e., $T_{ii} = 1-\epsilon$, $T_{ii}+T_{i,(i+1)_K}=1$,  $(i+1)_K := i \mod K + 1$. 
For instance-dependent label noise, the generation of noisy labels also depends on the features. We follow CORES \cite{cheng2020learning} to generate instance-dependent label noise. The generation process is detailed in Algorithm \ref{algorithm1}. With these definitions, \emph{asymm./inst. $\epsilon$} in Table \ref{hoc_estimator} denotes asymmetric/instance-dependent label noise with noise rate $\epsilon$.

\textbf{Model pre-training}. The network structures of all the three encoders in the Table \ref{hoc_estimator} are ResNet50 \cite{he2016deep}. Note that the encoders which generate features to estimate transition matrix can be pre-trained on different datasets. For example, HOC \cite{zhu2021clusterability} utilizes ImageNet pre-trained encoders to generate features for CIFAR. Thus, following the pipeline of disentangled features generation \cite{wang2021self}, we pre-train all the three encoders on CIFAR100 dataset and generate feature for CIFAR10 to estimate transition matrix. The first encoder is trained under 0.1 symmetric label noise rate to simulate the weakly-supervised features while the second and third encoder is trained via self-supervised learning (SSL). Recall the goal of SSL is to learn a good representation without accessing labels. 
In this paper, we adopt SimCLR \cite{chen2020simple} and IPIRM \cite{wang2021self} to perform SSL pre-training. SimCLR, as a representative work on SSL literature, learns a good represention based on InfoNCE loss \cite{van2018representation}. However, it is shown that the features learned by SimCLR are only \emph{partly} disentangled on some simple augmentation features such as rotation and colorization \cite{wang2021self}. Thus IPIRM proposes a learning algorithm that  embeds InfoNCE loss into IRM (Invariant Risk Minimization) framework \cite{arjovsky2019invariant} to learn \emph{fully} disentangled features. We train SimCLR model and IPIRM model by referring official codebase of IPIRM \footnote{https://github.com/Wangt-CN/IP-IRM}. The pre-trained models, as well as evaluation code are all released in the supplementary material. 
\textbf{Estimation error of Transition matrix}. 
After training these three encoders, we fix the encoder and generate features from raw samples to estimate the noise transition matrix using Global HOC estimator \cite{zhu2021clusterability}. The hyper-parameters for estimating transition matrix are consistent with official implementation of HOC \footnote{https://github.com/UCSC-REAL/HOC}: optimizer: Adam, learning rate: 0.1, number of iterations: 1500. After training, we evaluate the performance via absolute estimation error defined below:
\begin{equation*}
    \text{err} = \frac{\sum_{i=1}^{K}\sum_{j=1}^{K}|\hat{T}_{i,j} - T_{i,j}|}{K^{2}} * 100,
\end{equation*}
where $\hat{T}$ is the estimated noise transition matrix, $T$ is the real noise-transition matrix, $K$ is the number of classes in the dataset.

\begin{algorithm*}[!t]
	\caption{Instance-Dependent Label Noise Generation}
	\label{algorithm1}
	\begin{algorithmic}[1]
		\renewcommand{\algorithmicrequire}{\textbf{Input:}}
		\renewcommand{\algorithmicensure}{\textbf{Iteration:}}
		\REQUIRE ~~\\
		 1: Clean examples ${({x}_{n},y_{n})}_{n=1}^{N}$; Noise rate: $\varepsilon$; Size of feature: $1\times S$; Number of classes: $K$.
		\ENSURE ~~\\
		2: Sample instance flip rates $q_n$ from the truncated normal distribution $\mathcal{N}(\varepsilon, 0.1^{2}, [0, 1])$;\\
        3: Sample   $W \in \mathcal{R}^{S \times K}$ from the standard normal distribution $\mathcal{N}(0,1^{2})$;\\
		\textbf{for} $n = 1$ to $N$  \textbf{do} \\
		4: \qquad $p = {x}_{n} \cdot  W$   ~~~~ \text{// \small Generate instance dependent flip rates. The size of $p$ is $1\times K$.} 
		\\
		5: \qquad  $p_{y_{n}} = -\infty$   ~~~~~\text{// \small Only consider entries different from the true label}\\
		6: \qquad $p = q_{n} \cdot \text{softmax}(p) $ ~~~~ \text{// \small Let $q_n$ be the probability of getting a wrong label}
		\\
		7: \qquad $p_{y_{n}} = 1 - q_{n}$ ~~~~\text{// \small Keep clean w.p. $1 - q_{n}$} 
		\\
		8: \qquad Randomly choose a label from the label space as noisy label $\tilde{y}_{n}$ according to $p$;
		\\
		\textbf{end for}\\
		
		\renewcommand{\algorithmicensure}{\textbf{Output:}}
		\ENSURE ~~\\
		9: Noisy examples ${({x}_{i},\tilde{y}_{n})}_{n=1}^{N}$.
	\end{algorithmic}
\end{algorithm*}

{ \begin{table*}[!t]
		\caption{Comparison of test accuracy on CIFAR10 by using the estimated transition matrix. 
	    }
		\begin{center}
			\begin{tabular}{c|cccc} 
				\hline 
			  Methods & \emph{ inst. 0.3} & \emph{ inst. 0.4}
			  & \emph{ inst. 0.5}& \emph{ inst. 0.6}
			 \\ 
			 \hline\hline
			     FW (SimCLR) &66.61&65.82&64.51&62.81\\
			     FW (IPIRM) & 73.24&72.54&71.33&69.42\\
			     \hline
			\end{tabular}
		\end{center}
		\label{hoc_estimator_train}
	\end{table*}
}

{ \begin{table*}[!t]
		\caption{Comparison of test accuracy on CIFAR100 by using different DNN initialization. 
	    }
		\begin{center}
			\begin{tabular}{c|cccc} 
				\hline 
			  Methods & \emph{ inst. 0.3} & \emph{ inst. 0.4}
			  & \emph{ inst. 0.5}& \emph{ inst. 0.6}
			 \\ 
			 \hline\hline
				CE (random init) & 43.47&35.17 &27.07 &18.25 \\
			     CE (SimCLR init) &58.95 &49.7 &36.87 &25.07 \\
			     CE (IPIRM init) & 64.92 & 56.18&43.75 &30.36 \\
			     \hline
			\end{tabular}
		\end{center}
		\label{diff_init}
	\end{table*}
}

\subsection{Training performance using estimated transition matrix}

We can further use the estimated transition matrix to perform forward loss correction (FW) \cite{patrini2017making}. 
Table \ref{hoc_estimator_train} records the performance of FW by using the estimated transition matrix of SimCLR and IPIRM. The hyper-parameters for all the experiments in Table \ref{hoc_estimator_train} are the same: optimizer: SGD, training epochs: 100, learning rate: 0.1 for first 50 epochs and 0.01 for last 50 epochs, batch-size: 256. From the results, we can observe that the test accuracy increases as features become more disentangled.

\subsection{Initializing DNN using disentangled features}

Except for estimating transition matrix, we can directly use disentangled features to perform training on noisy dataset. Table \ref{diff_init} shows the effect of using disentangled features as DNN initialization on CIFAR100. The hyper-parameters for all the experiments in Table \ref{diff_init} are consistent with Table \ref{hoc_estimator_train}. From the results, 
We can observe that even with vanilla Cross Entropy loss, the disentangled features are still beneficial to the performance.

\end{document}